\documentclass[letterpaper]{article} 
\usepackage{aaai25}  
\usepackage{times}  
\usepackage{helvet}  
\usepackage{courier}  
\usepackage[hyphens]{url}  
\usepackage{graphicx} 
\usepackage{natbib}  
\usepackage{caption} 
\usepackage{algorithm}
\usepackage{algorithmic}

\usepackage{newfloat}
\usepackage{listings}

\usepackage{amsmath}
\usepackage{amsfonts}

\DeclareCaptionStyle{ruled}{labelfont=normalfont,labelsep=colon,strut=off} 
\floatstyle{ruled}
\newfloat{listing}{tb}{lst}{}
\floatname{listing}{Listing}
\title{GQWformer: A Quantum-based Transformer for Graph Representation Learning}
\author{
    Lei Yu,
    Hongyang Chen\thanks{Corresponding Auther.},
    Jingsong Lv,
    Linyao Yang
}
\affiliations{
    Research Center for Data Hub and Security, Zhejiang Lab\\
    yu1lei@mail.ustc.edu.cn, \{hongyang, jingsonglv, yangly\}@zhejianglab.com
}

\usepackage{bibentry}

\begin{document}

\maketitle

\begin{abstract} 
Graph Transformers (GTs) have demonstrated significant advantages in graph representation learning through their global attention mechanisms. However, the self-attention mechanism in GTs tends to neglect the inductive biases inherent in graph structures, making it chanllenging to effectively capture essential structural information.  
To address this issue, we propose a novel approach that integrate graph inductive bias into self-attention mechanisms by leveraging quantum technology for structural encoding. In this paper, we introduce the Graph Quantum Walk Transformer (GQWformer), a groundbreaking GNN framework that utilizes quantum walks on attributed graphs to generate node quantum states. These quantum states encapsulate rich structural attributes and serve as inductive biases for the transformer, thereby enabling the generation of more meaningful attention scores. By subsequently incorporating a recurrent neural network, our design amplifies the model's ability to focus on both local and global information. 
We conducted comprehensive experiments across five publicly available datasets to evaluate the effectiveness of our model. These results clearly indicate that GQWformer outperforms existing state-of-the-art graph classification algorithms. 
These findings highlight the significant potential of integrating quantum computing methodologies with traditional GNNs to advance the field of graph representation learning, providing a promising direction for future research and applications.
\end{abstract}

\section{Introduction}
Inspired by the celebrated success of Transformers in modeling structured data across various domains, researchers have begun to explore the application of Transformers to graph data \cite{rong2020self}. Unlike traditional attention networks, which aggregate node features solely from adjacent nodes, Graph Transformers (GTs) consider pairwise relationships between all nodes within a graph. This approach allows each node to directly attend to every other node, thereby capturing long-range dependencies and interactions from distant nodes, significantly enhancing their expressive power.

\begin{figure}[t]
\centering
\includegraphics[width=0.4\textwidth]{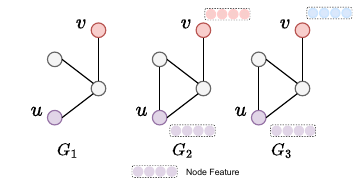} 
\caption{The comparison of encoding strategies. In graphs $G_1$ and $G_2$, the positional encodings between nodes $u$ and $v$ based on shortest paths are identical, yet their structural encodings differ due to distinct structures. Conversely, in $G_2$ and $G_3$, their structural encodings are identical when considering only the pure structural information. However, if the node attributes or edge weights differ, their attribute-aware structural encodings should vary.}
\label{fig1}
\end{figure}

However, the self-attention mechanism in GTs often neglects the inherent inductive biases of graphs, especially those related to graph topology.  
The absence of structural information consideration leads to indiscriminate attention to all nodes, thereby failing to discern the instrinsic differences within the graph. To mitigate this, numerous studies have centered on encoding positional or structural information to model these inductive biases, thereby bolstering the Transformer’s capacity to handle graph data \cite{ying2021transformers, chen2022structure, yeh2023random}. While significant strides have been made, the full potential of integrating graph structure information with node features remains untapped. Notably, as illustrated in Figure \ref{fig1}, for structurally identical graphs with different features, their attribute-aware structural encodings should vary, but existing methods fail to model this differentiation. In addition, the over-globalizing problem in GTs \cite{xing2024less} indicates that attention patterns, when indiscriminately augmented with globalizing properties, can lead to the accumulation of redundant information from distant nodes. Therefore, it is imperative to focus more on the local nuances within the graph. 

In order to resolve the aforementioned issues in GTs, we propose Graph Quantum Walk Transformer (GQWformer), an innovative framework that integrates GTs with intrinsic inductive bias by leveraging Quantum Walks (QWs) to encode the structural information of graphs. Specifically, GQWformer comprises two major components: the Graph Quantum Walk Self-attention Module (GQW-Attn) and the Graph Quantum Walk Recurrent Module (GQW-Recu). 
Our approach begins with the execution of attribute-aware QWs on a node-attributed graph. This encoding approach is sentitive to both the topological and contextual information of the graph, capturing a comprehensive representation of the graph’s structural attributes. The learned QW encodings are then utilized by GQW-Attn to measure the correlations between node pairs. By directly incorporating QW encodings into the attention biases, GQW-Attn ensures precise control over each node's attention weights relative to other nodes. This integration of QW encodings effectively embeds the graph's structural attributes into the GT model. 
Subsequently, GQW-Recu processes the complete sequences drived from the QWs, preserving and effectively utilizing the temporal and sequential dependencies to learn robust embedding for target nodes, which enhances the local processing power of the model. 
Additionally, the length of the QWs dictates the extent of each node's interactions, allowing for a flexible adjustment of the model’s receptive field. Importantly, the QWs are learnable, meaning they can dynamically adjust the encodings based on the specific characteristics of the graph data over time. This learnability ensures continuous optimization of QW encodings, providing a tailored inductive bias that effectively guides the learning process. As a result, the introduced QW encoding, rooted in the graph structural bias, significantly enhances the model's performance in graph learning tasks. Our contributions are summarized as follows.

\begin{itemize}
\item Addressing the general attributed node embedding problem, we propose a novel and potent framework-GQWformer, this tailored graph network harnesses the global structural information of the graph while distinguishing local information. It achieves this by ingeniously merging the advantages of Transformer and recurrent networks with QWs.
\item We use QW encoding to simultaneously extract topological and contextual information between pairs of nodes, leveraging the unique properties of QWs. This novel approach provides a comprehensive and nuanced understanding of the relationships within the graph.
\item We validate the effectiveness of our approach through empirical testing on five public benchmark datasets for graph classification where both network structure and node attributes information can be modeled. Our model exhibits a competitive performance advantage over RWC \cite{yeh2023random}.
\end{itemize}

\section{Related Work}
\subsubsection{Graph Transformers.} The transformer architecture have been applied to graph modeling, leading to the proposal of various encoding strategies. These strategies include laplacian eigenfunctions \cite{kreuzer2021rethinking}, node degree centrality \cite{ying2021transformers}, kernel distance \cite{mialon2021graphit}, shortest paths \cite{ying2021transformers}, random walks \cite{yeh2023random} and structure-aware methods \cite{chen2022structure}. Additionally, The issue of over-globalizing in GTs suggests the necessity of integrating local modules. Consequently, multiple studies \cite{zhang2022hierarchical, wu2022nodeformer, liu2023gapformer, yeh2023random, kong2023goat, wu2024simplifying} have been conducted to capture local information either implicitly or explicitly.

\subsubsection{Quantum Graph Learning.} Recently, quantum computing, an emerging research domain, has demonstrated substantial potential in machine learning, including graph learning \cite{liu2021rigorous, tang2022quantum}. With unique principles like superposition and entanglement, quantum computing redefines traditional paradigms of information understanding and processing. When applied to graph data, these principles facilitate effective graph feature characterization within a high-dimensional Hilbert space \cite{cong2019quantum, schuld2019quantum}, enabling the extraction of undetectable and atypical graph patterns \cite{huang2021power}.
Indeed, quantum computing offers a novel paradigm for handling graph-structured data, thereby revolutionizing graph learning \cite{yu2023quantum}. Quantum walks (QWs), the quantum equivalent of their classical counterparts, serve as a universal model for quantum computing. These walks provide a powerful method for encoding graph data using qubits and representing them with quantum states, proving effective in graph learning \cite{zhang2019quantum, bai2021learning, yan2022towards}.

\section{Preliminary}
In this paper, we focus on node-attributed graphs. 
Let $G=(\mathbf{V}, \mathbf{E})$ denote a graph with an adjacency matrix $\mathbf{A}\in \{0, 1\}^{n\times n}$ and a node feature matrix $\mathbf{X}\in \mathbb{R}^{n\times d}$. In the following, Scalars or elements of a set are denoted by italic lowercase letters, vectors are denoted by boldface lowercase letters and matrices are represented by boldface captical letters.
\subsection{Transformer}
Recent advancements have leveraged the Transformer architecture \cite{vaswani2017attention} to aggregate node features using the self-attention mechanism. 
The self-attention mechanism is mathematically defined as follows:
\begin{equation}
\begin{aligned}
\mathbf{a}_i^l &=\frac{\sum_{v_j\in \mathbf{V}}exp(\mathbf{a}_i^{(l-1)\top}\mathbf{k}_j^{(l-1)})\mathbf{v}_j^{(l-1)}}{\sum_{v_j\in \mathbf{V}}exp(\mathbf{q}_i^{(l-1)\top}\mathbf{k}_j^{(l-1)})}, \\
\mathbf{h}_i^l &=\mathbf{a}_i^l+\gamma \mathbf{h}_i^{(l-1)}.
\end{aligned}
\label{eq:transformer}
\end{equation}
Here, the node embedding of $v_i$ at the $l$-th layer, denoted as $\mathbf{h}_i^l$, is aggregated from the embeddings of all nodes in the graph. The term $\mathbf{h}_i^{(l-1)}$ represents the node embedding from the previous layer. To compute the self-attention, each node embedding $\mathbf{h}_i^{(l-1)}$ is projected into three distinct vectors: the query vector $\mathbf{q}_i^{(l-1)}$, the key vector $\mathbf{k}_i^{(l-1)}$, and the value vector $\mathbf{v}_i^{(l-1)}$, using the projection matrices $\mathbf{W}^Q$, $\mathbf{W}^K$ and $\mathbf{W}^V$, respectively. The resultant attention vector $\mathbf{a}_i^l$ is then combined with the previous layer's embedding 
$\mathbf{h}_i^{(l-1)}$, scaled by a factor $\gamma$, to form the new node embedding $\mathbf{h}_i^l$. This process enables the model to capture intricate relationships and dependencies across the entire graph by allowing each node to weigh the influence of every other node. 

\subsection{Quantum Computing}
In quantum computing, the basic unit of information is a qubit. Two possible states for a qubit are the computational basis states $|0\rangle=(1,0)$ and $|1\rangle=(0,1)$. Notation like '$| \rangle$' is called bra-ket notation. A qubit is also possible to form linear combinations of states, often called superpositions:
\begin{align}
|\psi\rangle = \alpha|0\rangle + \beta|1\rangle, \label{eq:quantum_state}
\end{align}
where the number $\alpha$ and $\beta$ are complex numbers that satisfy $\|\alpha\|^2 + \|\beta\|^2=1$. Formally, a quantum system on $n$ qubits lives in the $n$-fold tensor product Hilbert space $\mathcal{H}=(\mathbb{C})^{\otimes d}$ with resulting dimension $2^d$. A quantum state is represented as a unit vector $|\psi\rangle\in \mathcal{H}$. A quantum gate is a unitary operation $\mathbf{U}$ on $\mathcal{H}$, and the action of a unitary operator is denoted as $|\psi_2\rangle=\mathbf{U}|\psi_1\rangle$. The state space of a composite system is the tensor product of the state space of its components.

\subsection{Graph Quantum Walk}
Quantum walks (QWs), introduced by \cite{aharonov1993quantum}, represent a quantum analog of classical random walks. Unlike the stochastic evolution seen in classical random walks, QWs evolve through a unitary process.  
This unitary evolution gives rise to fundamentally different behaviors compared to classical walks, primarily due to the phenomenon of interference between different trajectories of the walker. Such interference can lead to faster spreading and different probability distributions, which are key features exploited in various quantum algorithms. Two kinds of QWs have been introduced in the literature, namely, continuous time QWs \cite{farhi1998quantum} and discrete time QWs \cite{ambainis2003quantum}. In this paper, we adopt the discrete time QW framework for general graphs as outlined in \cite{kendon2006quantum}.

Given a graph $G(\mathbf{V}, \mathbf{E})$, where $\mathbf{V}$ and $\mathbf{E}$ are vertex set and edge set respectively. The formulation of QWs involves defining two crucial Hilbert spaces. One is the position Hilbert space $\mathcal{H}_p$ which encapsulates the superposition over various positions, i.e., nodes, in the graph. Formally, $\mathcal{H}_p$ is defined as the span of the position basis vector $\{|v\rangle, v\in \mathbf{V}\}$. The position vector of a QW can then be expressed as a linear combination of position state basis vectors:
\begin{align}
|\psi_p\rangle = \sum_{v\in V} \alpha_v |v\rangle, \label{eq:position_state}
\end{align}
where $\{\alpha_v, v\in \mathbf{V}\}$ are the complex amplitudes satisfying the unit $L_2$-norm condition $\begin{matrix} \sum_{v} \|\alpha_v\|^2  = 1 \end{matrix}$, $\|\alpha_v\|^2$ is the probability of finding the walker at vertex $v$. This formulation allows for the QW to be in a superposition of multiple vertices simultaneously, a fundamental aspect of quantum mechanics that distinguishes QWs from their classical counterparts.

Another one is the coin Hilbert space $\mathcal{H}_c$ capturing the superposition of possible directions in which the walker can move from each node. Formally, $\mathcal{H}_c$ is defined as the span of the coin basis vectors $\{|i\rangle, i\in \left[1,..., d\right]\}$, where $i$ indexes the edges incident to a vertex $v$ and $d$ is the maximum degree of the graph. The coin state of a QW at vertex $v$ can then be expressed as a linear combination of coin state basis vectors:
\begin{align}
|\psi_c\rangle = \sum_{i\in \left[1,..., d\right]} \beta_{v,i}|i\rangle, \label{eq:coin_state}
\end{align}
where the coefficients $\{\beta_{v,i}, i\in \left[1,..., d\right]\}$ satisfy the unit $L_2$-norm condition $\begin{matrix} \sum_{i} \|\beta_{v,i}\|^2  = 1 \end{matrix}$. If a measurement is performed on the coin state of the walker at vertex $v$, the quantity $\|\beta_{v,i}\|^2$ represents the probability of observing the walker
at node $v$ and selecting the $i$-th neighbor of $v$ for its subsequent move. Overall, the comprehensive state space of the QW is described as the tensor product of the position and coin Hilbert spaces $\mathcal{H} = \mathcal{H}_p \otimes \mathcal{H}_c$.

\begin{figure*}[t]
\centering
\includegraphics[width=1.0\textwidth]{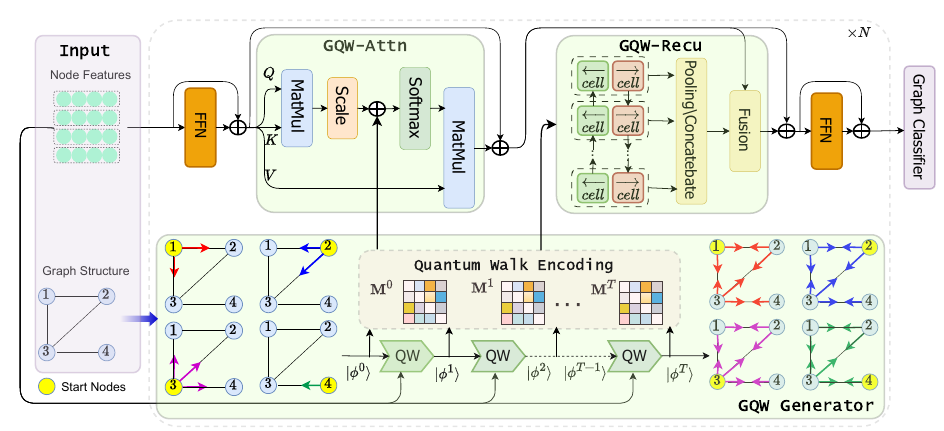} 
\caption{GQWformer. The model is composed of several GQWformer blocks. Each block consists of a GQW Generator, a Feed Forward Module (FFN), a GQW-Attn, a GQW-Recu, and a second FFN. With QW encoding, the structural information is effectively injected into the GQWformer.}
\label{fig2}
\end{figure*}

The time-evolution of QW over a graph is governed by two key unitary operators: the coin operator and the shift operator. Let $|\phi^t\rangle = |\psi_p^t\rangle \otimes |\psi_c^t\rangle$ denote the state of the walker at time $t$. A single step in the QW process  involves the sequential application of these two operations. Firstly, the coin operator $\mathbf{C}$ is applied, which operates exclusively on the coin Hilbert space, effecting a unitary transformation of the coin state at each vertex. This transformation can be expressed as:
\begin{align}
|\psi_p^t\rangle \otimes |\psi_c^{t+1}\rangle = (\mathbf{I} \otimes \mathbf{C})(|\psi_p^t\rangle \otimes |\psi_c^t\rangle), \label{eq:coin_operator}
\end{align}
where $\mathbf{I}$ denotes the identity operator. Subsequently, a unitary shift operator $\mathbf{S}$ is applied to swap the states of vertices connected by edges, thus facilitating the walker’s movement across the graph. The shift operator acts on both the coin and position Hilbert spaces. Formally, this step can be described as:
\begin{align}
|\phi^{t+1}\rangle = |\psi_p^{t+1}\rangle \otimes |\psi_c^{t+1}\rangle = \mathbf{S}(|\psi_p^t\rangle \otimes |\psi_c^{t+1}\rangle). \label{eq:shift_operator}
\end{align}
In shorthand notation, the unitary evolution of the walk is governed by the composite operator $\mathbf{U} = \mathbf{S}(\mathbf{I}\otimes \mathbf{C})$. If the initial state of the QW on the graph is denoted by $|\phi^0\rangle$, then after $t$ time steps, the state of the walks is described by $|\phi^t\rangle = \mathbf{U}^t|\phi^0\rangle$.

It's essential to note that QWs diverge significantly from classical random walks, primarily due to the pronounced influence of the initial superposition and the coin operator. These elements introduce additional degrees of freedom, enabling deep learning techniques to more effectively fit data through a controlled diffusion process.

\section{Method}
This subsection presents the methodology of GQWformer, designed to capture global structural information while emphasizing local details within graphs. The central concept of this approach is to harness the power of QWs on attributed graphs to model and interpret the intricate interactions between nodes, thereby producing highly informative encodings. Figure \ref{fig2} provides an illustrative overview of the GQWformer architecture, We will now delve into the specifics of each component.

\subsection{Attribute-aware Graph Quantum Walk}
We employ the framework of multiple non-interacting QWs on arbitrary graphs, as introduced in \cite{rohde2011multi}, to facilitate the learning of structural encodings in graph data.

Consider a graph with $n$ vertices, the QW process involves $n$ separate, non interacting walks running in parallel, with each walk originating from a unique node in the graph. Suppose the maximum degree of the graph is $d$, the initial state is encapsulated in the superposition tensor $|\phi^0\rangle \in \mathbb{C}^{n\times n\times d}$, which represents the collective state of the $n$ walkers.
The evolution of the QW can be decomposed into a sequence of discrete steps.
At each step $t$, the current superposition tensor $|\phi^t\rangle$ is updated using a set of coin operators, following by swapping the states along the edges of the graph.

As previously mentioned, coin operators are pivotal in modifying the spin state of the QW, thereby governing the evolution of this non-classical walk process over the graph. These coin operators can exhibit spatial variability across different nodes in the graph or temporal variability along the steps of the walk. However, the basic QW framework does not inherently account for node feature information. To address this, we can learn a function that generates distinct coin operations at each node, based on the features of neighboring nodes. This ensures that even structurally identical graphs can produce different walks if their node features differ. Consequently, learning QWs on a graph is achieved by learning the appropriate coin operators. In general, the function that generates the coin operators could be an arbitrary function that yields a valid coin operator $\mathbf{C}\in \mathbb{C}^{d\times d}$.


In this paper, we focus on the use of elementary unitary matrices for the coin operators. These matrices take the form: 
\begin{align}
\mathbf{U}=\mathbf{I}-2\mathbf{e}\mathbf{e}^\top/(\mathbf{e}^\top \mathbf{e}), \label{eq:matrix_form}
\end{align}
where $\mathbf{I}$ represents the identity matrix and $\mathbf{e}$ is an arbitrary vector. This specific form of unitary matrix can be computed efficiently during the forward pass of the neural network and its gradients can similarily be computed efficiently during backpropagation, making it well-suited for integration into learning algorithms. For a given node $v_i$, the coin operator can be generated as follows:
\begin{align}
\mathbf{C}_i=\mathbf{I}-2g(v_i)g(v_i)^\top/(g(v_i)^\top g(v_i)). \label{eq:coin_form}
\end{align}
Here, $g(v_i)$ is a function related to the features of node $v_i$. This approach ensures that the coin operators are adaptively learned based on the node features, thereby enabling the QW process to capture both structural and feature-based information within the graph.

Inspired by graph attention networks and diverging from the approach presented in \cite{dernbach2019quantum}, we propose a novel function $g(v_i)$ designed to compute attention scores between the node $v_i$ and each of its neighbors. The function is designed as:
\begin{align}
g(v_i)=a(\mathbf{W}\mathbf{X}_{\mathcal{N}(v_i)}, \mathbf{W}\mathbf{X}_{d}^i), \label{eq:gen_func}
\end{align}
where $\mathbf{X}_{\mathcal{N}(v_i)}$ denotes the features matrix of the neighbors of $v_i$, $\mathbf{X}_{d}^i$ is a matrix with $d$ rows, and each row is the feature vector of $v_i$ itself, $\mathbf{W}\in \mathbb{C}^{F\times F'}$ is a learnable weight matrix that linearly transforms the feature vectors of nodes, and $a$ is an attention function that computes the attention scores. By incorporating these elements, our proposed function $g(v_i)$
can dynamically evaluate the influence of each neighbor and modify the spin states of the QW to prioritize specific neighbors.

A $T$-step QW generates a sequence of superposition states $\{|\phi^0\rangle, |\phi^1\rangle,..., |\phi^T\rangle\}$. Now this sequence captures complex patterns and relationships within the graph, making it a powerful tool for tasks such as graph classification.

By summing the squares of the spin states of each superposition state, we derive a sequence of matrices $\{\mathbf{M}^0, \mathbf{M}^1,..., \mathbf{M}^T\}$, where each $\mathbf{M}^i\in \mathbb{C}^{n\times n}$ for $i\in [1,..., T]$, we then take the sequence as the input for GQWformer, as will be introduced in the following section.

\subsection{Graph Quantum Walk Transformer}
While Transformer excel at modeling the global information, they often struggle to capture essential graph structural information. Here we extract structural infromation using quantum method. Building upon this foundation, we introduce GQWformer, which extends the original GT architecture by incorporating these QW encodings, thereby introducing a potent inductive bias into the self-attention mechanism. Recognizing that the GQW-Attn module may overlook certain local structure information, we further introduce the GQW-Recu module to supplement the focus on local structural details. 

\subsubsection{Graph Quantum Walk Self-attention Module} Compared to conventional GNNs, Transformer offers a more generalized framework by meticulously analyzing the pairwise correlations between every node. One of the significant advantages of Transformers is their global receptive field, which allows each node to attend to the information from any other node in the graph, thus enabling a comprehensive processing of node representations. However, the self-attention mechanism in Transformers primarily focuses on calculating the semantic similarity between nodes, often neglecting the crucial topological information inherent in the graph. Therefore, to fully harness the power of Transformers in graph-based tasks, it is imperative to effectively integrate the structural information of graphs into the model.

\begin{figure}[t]
\centering
\includegraphics[width=0.5\textwidth]{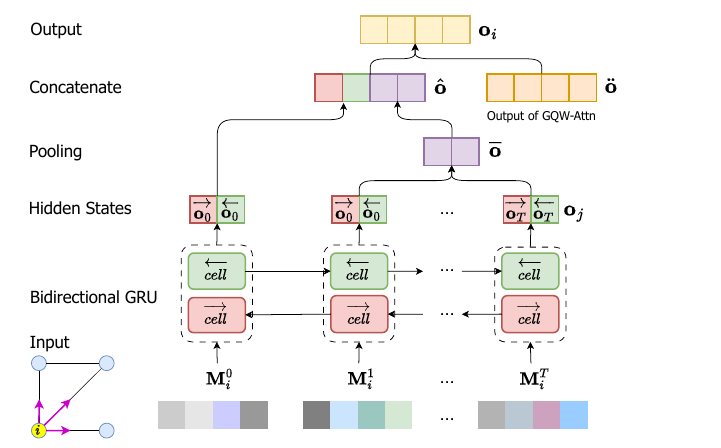} 
\caption{The architecture of the Graph Quantum Walk Recurrent Module}
\label{fig3}
\end{figure}

With the sequence $\{\mathbf{M}^0, \mathbf{M}^1,..., \mathbf{M}^T\}$ obtained in the previous section, we interpret $\mathbf{M}^T$ as a structural encoding matrix, where the $(i,j)$-th element of $\mathbf{M}^T$, denoted as $p_{ij}$, encodes the distance between node $v_i$ and node $v_j$. As previous described, this distance is a trainable parameter that is sensitive to both the structural and attribute information of the graph. The proposed QW encoding serves as a bias term in the attention module, thereby incorporating an attention bias into the attention score calculation. Specifically, Graph Quantum Walk Self-attention Module (GQW-Attn) is formulated as:
\begin{align}
\mathbf{a}_i^l&=\frac{\sum_{v_j\in V}exp(\mathbf{q}_i^{(l-1)\top}\mathbf{k}_j^{(l-1)}+p_{ij})\mathbf{v}_j^{(l-1)}}{\sum_{v_j\in V}exp(\mathbf{q}_i^{(l-1)\top}\mathbf{k}_j^{(l-1)}+p_{ij})}. \label{eq:gqw-attn}
\end{align}
By incorporating the QW encoding as an attention bias, the GQW-Attn ensures that the attention mechanism not only captures the semantic similarity but also respects the topological structure of the graph, leading to more expressive and robust node representations.

As the parameter $T$ increases, the QW extends its exploration to larger regions of the graph. Consequently, $\mathbf{M}^T$ encodes more valuable information between node pairs compared to the initial state. This extended exploration allows the model to capture long-range dependencies and intricate structural relationships within the graph. The ability to dynamically adjust the extent of this exploration via the parameter $T$ provides the model with flexibility and adaptability to various graph structures and scales.

\subsubsection{Graph Quantum Walk Recurrent Module}
The sequence $\{\mathbf{M}^0, \mathbf{M}^1,..., \mathbf{M}^T\}$ encapsulates the interaction of nodes with their neighbors, dictated by the graph's topological structure and node attributes. Recurrent neural networks \cite{schuster1997bidirectional} form the foundation of many sequence-to-sequence methods, the work of \cite{huang2019graph} extends this paradigm to graph domains by introducing graph recurrent networks (GRN), where GCN can be interpreted as a special case of GRN. In this vein, we explore to utilize GRN to learn the quantum embedding for the initial nodes of QWs.

\begin{algorithm}[tb]
\caption{Graph Quantum Walk Transformer (GQWformer)}
\label{alg:algorithm}
\begin{algorithmic}[1] 
\STATE Adding degree encoding to node features.
\FOR {$k\leftarrow 1$ to $K$}
\STATE  Perform QW to produces a sequence of encoding matrix $\{\mathbf{M}^0, \mathbf{M}^1,..., \mathbf{M}^T\}.$
\STATE $\mathbf{H}_F^{k-1}=\mathbf{H}^{k-1}+\frac{1}{2}\mathrm{FFN}(\mathbf{H}^{k-1})$
\STATE $\mathbf{H}_A^{k-1}=\mathbf{H}_F^{k-1}+\mathrm{GQW\text{-}Attn}(\mathbf{H}_F^{k-1},\mathbf{M}^T)$
\STATE $\mathbf{H}_R^{k-1}=\mathbf{H}_A^{k-1}+\mathrm{GQW\text{-}Recu}(\mathbf{H}_A^{k-1},\mathbf{M}^0,..., \mathbf{M}^T)$
\STATE $\mathbf{H}^{k}=\mathbf{H}_R^{k-1}+\frac{1}{2}\mathrm{FFN}(\mathbf{H}_R^{k-1})$
\ENDFOR
\STATE \textbf{return} the final embedding of the virtual node
\end{algorithmic}
\end{algorithm}

Specifically, we take the quantum encoding sequence $\{\mathbf{M}^0, \mathbf{M}^1,..., \mathbf{M}^T\}$ as the primary input. As depicted in Figure \ref{fig3}, the first input should be $\mathbf{M}_i^0$, which are the initial state of the first node, specifically the target node. To precess this sequence, we employ two sequences of GRU cells, they take the same input, but run in oppsite directions, one running forward and the other running backward. In this bidirectional GRU setup, $\overleftarrow{\mathbf{o}}_j$ denotes the learned representation for the $j$-th backward cell, while $\overrightarrow{\mathbf{o}}_j$ represents the corresponding forward cell. 
We first concatenate $\overrightarrow{\mathbf{o}}_j$ and $\overleftarrow{\mathbf{o}}_j$ to form the output of the layer, denoted as $\mathbf{o}_j$. Notably, $\mathbf{o}_0$ is the output of the target node, it already combines the information in the sequence. Subsequently, we apply a pooling operation to merge the outputs $\{\mathbf{o}_1,...,\mathbf{o}_T\}$ into a single vector $\overline{\mathbf{o}}$. This pooling step serves to further distill the information contained within the entire sequence. Then we obtain $\hat{\mathbf{o}}$ by concatenating $\mathbf{o}_0$ with $\overline{\mathbf{o}}$. Assuming the output of GQW-Attn is $\ddot{\mathbf{o}}$, the final embedding of the target nodes is defined as $\mathbf{o}_i=\hat{\mathbf{o}} + \ddot{\mathbf{o}}$.

The pooling operation within GQW-Recu serves to inductively learn from the neighboring nodes. Moreover, GQW-Recu capitalizes on the inherent order information of the sequence, ensuring that temporal and directional dependencies are thoroughly integrated into the node representations. As a result, GQW-Recu yield a comprehensive and more informative representation of the target nodes.

The integration of GQW-Recu ensures that our model is not only aware of global structures but also deeply attuned to local intricacies, which are vital for accurate graph analysis.

\subsubsection{Readout}
Following the methodology outlined in \cite{yeh2023random}, we add a special node to the graph, and make it connect to all other nodes within the graph. This virtual node facilitates efficient information exchange among nodes, thereby enhancing overall performance \cite{gilmer2017neural}. Furthermore, since the virtual node aggregates information from all nodes in the graph, we adopt its hidden feature as the whole graph embedding and train an additional classifier for the downstream tasks.

\section{Experiments}
In this section, we conduct a comprehensive empirical analysis of the proposed GQWformer framework, focusing on its performance in graph classification tasks. We evaluate GQWformer using five real-world network datasets, benchmarking its effectiveness against several state-of-the-art GNNs. 

\begin{table*}[t]
\centering

\begin{tabular}{l c c c c c}
\hline
\rule{0pt}{2.5ex}Method & MUTAG$\uparrow$ & PTC$\uparrow$ & PROTEINS$\uparrow$ & IMDB-B$\uparrow$ & IMDB-M$\uparrow$ \\
\hline
\rule{0pt}{2.5ex}DGCNN \cite{zhang2018end} & $85.5\pm1.8$ & $58.6\pm2.5$ & $75.5\pm0.9$ & $70.0\pm0.9$ & $47.8\pm0.9$ \\
IGN \cite{maron2018invariant} & $83.9\pm13.0$ & $58.5\pm6.9$ & $76.6\pm5.5$ & $72.0\pm5.5$ & $48.7\pm3.4$ \\
GIN \cite{xu2018powerful} & $89.4\pm5.6$ & $64.6\pm7.0$ & $76.2\pm2.8$ & $75.1\pm5.1$ & $52.3\pm2.8$ \\
PPGNS \cite{maron2019provably} & $90.6\pm8.7$ & $66.2\pm6.6$ & $77.2\pm4.7$ & $73.0\pm5.8$ & $50.5\pm3.6$ \\
Natural GN \cite{de2020natural} & $89.4\pm1.6$ & $66.8\pm1.7$ & $71.7\pm1.0$ & $73.5\pm2.0$ & $51.3\pm1.5$ \\
RWNN \cite{nikolentzos2020random} & $89.2\pm4.3$ & $65.8\pm5.5$ & $74.7\pm3.3$ & $70.8\pm4.8$ & $48.8\pm2.9$ \\
CRaWl \cite{toenshoff2021graph} & $88.2\pm5.6$ & $63.9\pm4.9$ & $74.1\pm4.4$ & $72.7\pm2.8$ & $47.8\pm3.9$ \\
CIN \cite{bodnar2021weisfeiler} & $92.7\pm6.1$ & $68.2\pm5.6$ & $77.0\pm4.3$ & $75.6\pm3.7$ & $52.7\pm3.1$ \\
GSN \cite{bouritsas2022improving} & $92.2\pm7.5$ & $68.2\pm7.2$ & $76.6\pm5.0$ & $77.8\pm3.3$ & $54.3\pm3.3$ \\
GNN-AK \cite{zhao2022from} & $91.3\pm7.0$ & $67.7\pm8.8$ & $77.1\pm5.7$ & $75.0\pm4.2$ & $52.7\pm4.8$ \\
RWC \cite{yeh2023random} & $94.7\pm3.7$ & $74.4\pm4.4$ & $79.4\pm4.7$ & $78.8\pm3.1$ & $54.5\pm2.9$ \\
\hline
\rule{0pt}{2.5ex}GQWformer (ours) & $\mathbf{95.2\pm3.0}$ & $\mathbf{76.7\pm4.2}$ & $\mathbf{80.7\pm4.3}$ & $\mathbf{79.3\pm1.7}$ & $\mathbf{55.3\pm2.1}$ \\
\hline
\end{tabular}
\caption{Results on graph classification}
\label{table2}
\end{table*}

\subsection{Experimental Settings}
\subsubsection{Datasets.} For graph classification, we test GQWformer on five TUDataset benchmarks \cite{morris2020tudataset} from various domains, including one biology (i.e., PROTEINS), two chemistry (i.e., MUTAG and PTC), and two social (i.e., IMDB-B and IMDB-M) datsets. 






\subsubsection{Baselines.} To demonstrate the effectiveness of our proposed method, we compare GQWformer with the following $11$ baselines: DGCNN \cite{zhang2018end}, IGN \cite{maron2018invariant}, GIN \cite{xu2018powerful}, PPGNS \cite{maron2019provably}, Natural GN \cite{de2020natural}, RWNN \cite{nikolentzos2020random}, CRaWl \cite{toenshoff2021graph}, CIN \cite{bodnar2021weisfeiler}, GSN \cite{bouritsas2022improving}, GNN-AK \cite{zhao2022from} and RWC \cite{yeh2023random}.

\subsubsection{Implementation Details.} In the task of graph classification, given the embedding representations of all graphs, and a set of training graphs with labels, the goal is to predict the labels of the remaining graphs. The classification performance is measured by the accuracy score.

We adhere to the conventional settings outlined in \cite{yeh2023random} to validate the performance of GQWformer. Specifically, for all experiments, we employ a linear learning rate scheduler with the end learning rate $1e-9$. The optimizer of GQWformer is AdamW with a weight decay of $0.01$. To mitigate the risk of exploding gradient, we set a gradient clipping value to $1$. Additionally, the dropout for FFN, GQW-Attn, and GQW-Recu modules is maintained at $0.1$. The QW is configured to a length of $4$, and the number of GQWformer blocks is also set to $4$.

In line with \cite{yeh2023random}, ten-fold cross-validation is employed to select the training and test sets. For each dataset, given the entire set of graphs $\mathcal{G}$, we randomly select $90\%$ of $\mathcal{G}$ as the training set, while the remaining is used as the test set.

\begin{table}[t]
\centering
\begin{tabular}{c c c c}
\hline
\rule{0pt}{2.5ex}GQW-Attn & GQW-Recu & Enc. & PTC Acc.$\uparrow$ \\
\hline
\rule{0pt}{2.5ex}$\checkmark$ & $\times$ & QW(ours) & 75.3 \\
$\times$ & $\checkmark$ & QW(ours) & 72.1 \\
$\checkmark$ & $\checkmark$ & vanilla QW & 72.1 \\
$\checkmark$ & $\checkmark$ & QW(inv) & 73.3 \\
$\checkmark$ & $\checkmark$ & QW(ours) & $\mathbf{76.7}$ \\
\hline
\end{tabular}
\caption{Ablation studies}
\label{table3}
\end{table} 

\begin{table}[t]
\centering
\begin{tabular}{l c c c c c c}
\hline
\rule{0pt}{2.5ex}Walk length & 3 & 4 & 5 & 6 & 7 & 8 \\
\hline
\rule{0pt}{2.5ex}PTC Acc.$\uparrow$ & 73.0 & $\mathbf{76.7}$ & 73.0 & 74.7 & 72.8 & 73.8 \\
\hline
\end{tabular}
\caption{Sensitivity test}
\label{table4}
\end{table} 

\subsection{Effectiveness Evaluation}
From Table \ref{table2}, it is evident that our method demonstrates performance on par with, and often superior to, various baseline methods. Specifically, for the chemistry datasets, GQWformer outperforms all baselines by at least $0.5\%$ on MUTAG, and $2.3\%$ on PTC in terms of accuracy. Similarly, GQWformer improves the accuracy by $1.3\%$ on the biological dataset PROTEINS, and for the social networks, it achieves an accuracy increase of $0.5\%$ on IMDB-B and $0.8\%$ on IMDB-M. Especially, RWC \cite{yeh2023random} is a recently proposed GNN model, which employs the RW-SAN to measure the global pairwise correlations and RW-Conv to carefully analyze the local substructures of graphs. Our GQWformer method consistently and significantly outperforms RWC on all five datasets by a large margin, underscoring its effectiveness and robustness in graph classification tasks. In summary, the empirical results presented in Table \ref{table2} clearly showcase the superiority of GQWformer over existing baseline methods, including the recent RWC model.

\subsection{Ablation Studies}
We conduct ablation studies to evaluate the importance of different modules in GQWformer on the PTC dataset, as presented in Table \ref{table3}. These studies provide a comprehensive analysis of each component's contribution to the overall performance of our model. 
Firstly, we validate the significance of the GQW-Attn module. The experimental results clearly demonstrate that incorporating GQW-Attn significantly enhances the performance of GQWformer. The adoption of GQW-Attn allows the model to effectively capture and utilize the structural information inherent in the graph data. Additionally, assigning attention based on QW further amplifies the model's performance, highlighting the crucial role of this mechanism in improving graph representation and ensuring a more accurate classification.
Next, we investigate the effectiveness of the GQW-Recu module. GQW-Recu leverages the inherent temporal and sequential dependencies of the QW sequence to focus on local details, thereby bolstering the performance of GQWformer. This module's ability to exploit these dependencies facilitates a more nuanced and comprehensive understanding of the graph's local structure, which translates into better classification accuracy. 
Furthermore, we compare different encoding strategies, including the vanilla QW, which operates independently of feature information, the invariant QW, which is also attribute-aware, as introduced in \cite{dernbach2019quantum}, and our proposed QW encoding, designed to be more sensitive to the graph's attribute information. Our encoding strategy proves to be more powerful than both the vanilla and invariant QW encodings on the PTC dataset. This superiority underscores the effectiveness of our approach in capturing the essential features and relationships within the graph data. In conclusion, the ablation studies highlight the indispensable roles of GQW-Attn and GQW-Recu in enhancing the performance of GQWformer. Additionally, our proposed QW encoding strategy demonstrates its effectiveness in graph classification tasks.

\subsection{Sensitivity Analysis}
In our study, a critical aspect of GQWformer is the evaluation of the impact of the QW length on model performance. To thoroughly investigate this, we conducted a series of experiments with varying walk lengths using the PTC dataset, as detailed in Table \ref{table4}. The GQWformer mechanism benefits from longer walk sequences by exploring a greater number of nodes within the graph, thereby potentially capturing more intricate global structural patterns. However, this increased exploration comes with inherent tradeoffs. Specifically, there is a balance to be struck between the enhanced performance derived from a more comprehensive exploration and the computational efficiency that may be compromised with longer walks. Additionally, there is a delicate equilibrium between the ability to capture global structural information versus local structural details. Our experimental results underscore the significance of walk length, demonstrating that it indeed has a substantial impact on model performance. Longer walks allow the model to capture more extensive global structures, but they also risk diluting the focus on local details and increasing computational costs. 
These findings highlight the necessity of carefully selecting the walk length to optimize the balance between performance and efficiency, as well as between global and local structural information.

\section{Conclusion}
We have proposed a novel GNN, termed Graph Quantum Walk Transformer (GQWformer), this innovative model leverages the unique properties of QWs to capture rich topological and contextual information from graph structures. By integrating a GQW-Attn module, GQWformer effectively extracts global pairwise relationships, while the GQW-Recu component adeptly focuses on local details. Experimental results demonstrate that GQWformer consistently achieves state-of-the-art performance across a diverse range of domains, including biological networks, chemical compounds, and social interaction graphs, particularly excelling in graph classification tasks.


\bibliography{aaai25}

\end{document}